\documentclass[10pt,twocolumn,letterpaper]{article}

\usepackage{cvpr}              

\usepackage[dvipsnames]{xcolor}

\definecolor{cvprblue}{rgb}{0.21,0.49,0.74}
\usepackage[pagebackref,breaklinks,colorlinks,citecolor=cvprblue]{hyperref}

\usepackage{graphicx}
\usepackage{amsmath}
\usepackage{amssymb}
\usepackage{booktabs}
\usepackage{multirow}
\usepackage{makecell}
\usepackage[pagebackref,breaklinks,colorlinks]{hyperref}
\usepackage{bm}
\usepackage{enumitem}
\usepackage[capitalize]{cleveref}

\usepackage{adjustbox}
\usepackage[accsupp]{axessibility} 

\usepackage{spverbatim}

\def\paperID{3} 
\def\confName{CVPR}
\def\confYear{2024}

\title{Improving face generation quality and prompt following with synthetic captions}

\author{Michail Tarasiou \qquad  Stylianos Moschoglou \qquad 
\qquad Jiankang Deng \qquad Stefanos Zafeiriou \\
Imperial College London 
}

\begin{document}
\maketitle
\begin{abstract}

Recent advancements in text-to-image generation using diffusion models have significantly improved the quality of generated images and expanded the ability to depict a wide range of objects. However, ensuring that these models adhere closely to the text prompts remains a considerable challenge. This issue is particularly pronounced when trying to generate photorealistic images of humans. Without significant prompt engineering efforts models often produce unrealistic images and typically fail to incorporate the full extent of the prompt information. This limitation can be largely attributed to the nature of captions accompanying the images used in training large scale diffusion models, which typically prioritize contextual information over details related to the person's appearance.
In this paper we address this issue by introducing a training-free pipeline designed to generate accurate appearance descriptions from images of people. We apply this method to create approximately 250,000 captions for publicly available face datasets. We then use these synthetic captions to fine-tune a text-to-image diffusion model. Our results demonstrate that this approach significantly improves the model’s ability to generate high-quality, realistic human faces and enhances adherence to the given prompts, compared to the baseline model.
We share our synthetic captions, pretrained checkpoints and training code \verb|https://github.com/michaeltrs/Text2Face|

\end{abstract}
    
\section{Introduction}
\label{sec:intro}

Recent years have witnessed a significant resurgence in generative modelling for a range of modalities such as text \cite{gpt2, instruct, llama}, vision \cite{dif1, dif2, dif3, dif5, latent_diffusion, dif4} and audio \cite{audio}. 

\begin{figure*}[t]
   \centering
   \includegraphics[trim={0 0 42.5cm 0}, clip, width=0.995\linewidth]{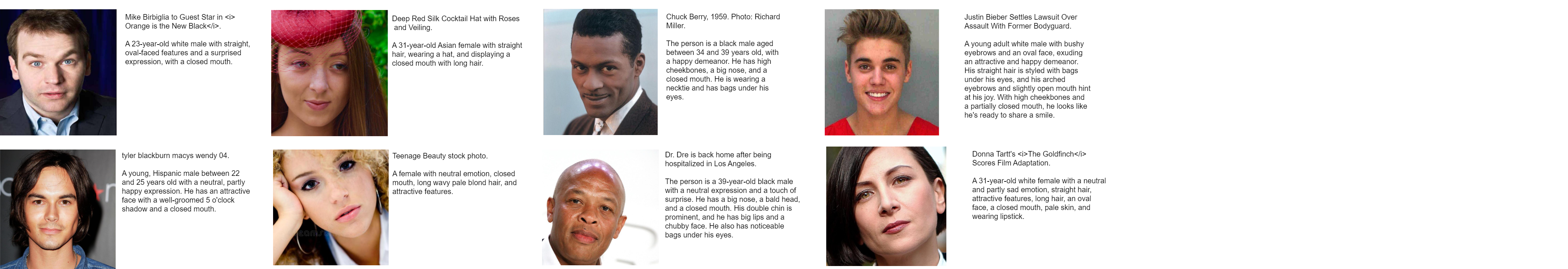}
   \caption{Comparison with captions from the LAION-Face dataset \cite{farl}. For each image we show the caption provided in LAION-Face (top) and the output of our synthetic captioning pipeline (bottom).}
    \label{fig:captions}
\end{figure*}

Particularly in vision, diffusion-based image generators have dramatically enhanced the quality and expressiveness of images generated from various input signals such as text, depth maps, and semantic labels just to name a few. 
Text-to-image models, trained on large-scale, internet-based image-caption pairs have demonstrated remarkable generative capabilities \cite{latent_diffusion, dif4}. However, captions describing images sourced from the internet often lack information that is useful for training such models. Although there is some signal to learn the content and style of images from their textual descriptions, most internet-based captions provide context for the image rather than accurately describing what is depicted. 
This discrepancy is particularly pronounced when working with images of humans, where internet-sourced captions can encompass multiple aspects such as the setting, profession or even the name of the person depicted, without necessarily describing their appearance. Some examples of image-caption pairs that exhibit these characteristics are shown in Fig.\ref{fig:captions}. As a result, ``in-the-wild'' data exhibit a low signal-to-noise ratio, which hinders effective training. Consequently, very large datasets are required. 
Furthermore, models trained in this manner often fall short of generating images that accurately follow detailed prompt descriptions. 

A solution to this problem can be found through utilizing curated datasets or synthetic captioning. 
Several studies have found that smaller, higher-quality annotations can lead to improved modeling outcomes  \cite{dalle3}. 
Building on this insight and utilizing the extensive literature and resources around face analysis, this paper proposes a pipeline for generating synthetic captions describing a person's appearance only from face images. 

Our contributions are twofold:
\begin{itemize}
    \item We propose a training-free captioning pipeline for face images based on extensive prior work in 2D and 3D face analysis. Specifically, we leverage a set of publicly available models, pre-trained on specific face analysis tasks (e.g., face and landmark detectors), to extract detailed appearance attributes. These attributes are far more detailed than the information typically contained in large-scale web captions. Then, we transform these outputs into word descriptions of attributes, producing a bag-of-words description for each face/image. These descriptions are then processed by a pretrained, instruction-tuned Large Language Model (LLM) to produce natural language image captions. We share the synthetic captions for approximately 250k face images from public datasets.
    \item Using our synthetic captions, we fine-tune a text-to-image diffusion model to produce high-quality, realistic face images that closely adhere to the provided text descriptions. We also make this model publicly available. 
\end{itemize}

\section{Synthetic Captions for Human Faces}\label{sec:method}
In this section, we present our training-free pipeline for obtaining textual descriptions of humans. Section \ref{sec:feature_extraction} describes the feature extraction process, specifically the pre-trained models used and their respective outputs. Section \ref{sec:llm} details how we use the extracted bag-of-words features to generate coherent textual descriptions.

\subsection{Feature extraction} \label{sec:feature_extraction}
We begin by applying a powerful {\bf face detection} model to all face images using a ResNet50 backbone from \cite{retinaface}. The output of the face detector provides the number of faces detected in each image, along with their locations and five facial landmarks for: the eye pupils, tip of the nose and left/right commissures of the mouth. We exclude all images containing more than one face and use the facial landmarks to crop square images around the face region. These crops are optionally aligned to predefined templates for inference with subsequent models. At this stage, we ensure that only the facial region is shown, with the face centered within the crop.

Next, we proceed with a second round of filtering aimed at removing images of animated humans and cartoons, which the face detector incorrectly identifies as humans. We have found {\bf CLIP} \cite{clip} to be an effective solution for identifying these images. We design CLIP text prompts to detect images either not depicting real humans or containing significant amounts of text overlaid over the depicted faces, such as album covers or film posters. Following this filtering, we use CLIP to assess if the person’s teeth or tongue are visible, as this information allows for better rendering, especially for teeth.

Subsequently, we proceed to feature extraction. First, we extract {\bf facial attributes} using \cite{farl} to obtain a list of attributes characterizing the person in the image. The pretrained model detects the presence of each of the following classes independently: 
\textsl{``5 o'clock shadow'', ``arched eyebrows'', ``attractive'', ``bags under eyes'', ``bald'', ``bangs'', ``big lips'',
``big nose'', ``black hair'', ``blond hair'', ``blurry'', ``brown hair'', ``bushy eyebrows'', ``chubby'',
``double chin'', ``eyeglasses'', ``goatee'', ``gray hair'', ``heavy makeup'', ``high cheekbones'', ``male'',
``mouth slightly open'', ``mustache'', ``narrow eyes'', ``no beard'', ``oval face'', ``pale skin'', ``pointy nose'',
``receding hairline'', ``rosy cheeks'', ``sideburns'', ``smiling'', ``straight hair'', ``wavy hair'',
``wearing earrings'', ``wearing hat'', ``wearing lipstick'', ``wearing necklace'', ``wearing necktie'', ``young''}.

We obtain a description of the emotions exhibited by the person through a {\bf emotion recognition} with single class categorical outputs using \cite{emotion}. Detected emotions are: \textsl{``anger'', ``disgust'', ``fear'', ``happiness'', ``sadness'', ``surprise'' and ``neutral''}. We retain the most dominant emotion detected. If no single emotion is clearly dominant, we retain the top two emotions and describe the person as exhibiting both.

Next, all images are processed for {\bf face parsing} using \cite{farl} resulting in dense class maps for all facial regions. We use these outputs to get a set of derivative attributes, such as the length of the hair and descriptions of the openness of the eyes and mouth.

Finally, we use a model to gather information about the {\bf age, gender, and race/ethnicity} of the person \cite{deepface2}.
We find that gender is predicted with a high degree of accuracy.
However, age is a subjective attribute that varies significantly among individuals, as people age at different rates. Therefore, using a specific age based on predictive models can be misleading. To address this, we treat age as a range rather than a precise value. This approach allows for more flexibility and realism in the synthetic labeling of human images. Specifically, we either sample an age from a distribution centered around the predicted age or categorize ages into five-year intervals to better accommodate the variability in aging among different people and the inaccuracies in model predictions.
For ethnicity, the model predicts the closest output among classes: \textsl{``black'', ``white'', ``asian'', ``middle eastern'', ``indian'', ``hispanic''}.

\subsection{Language Fusion} \label{sec:llm}
Following the completion of the feature extraction stage, we can associate each extracted feature with a descriptive word. Thus, we now have a set of features describing the person in an image in the form of a bag-of-words. To obtain a set of descriptions in the form of natural language, we simply use an instruct-based pre trained LLM, which we condition to output a description based on the provided bag of words. For this purpose, we utilise the Vicuna 1.5 13B model \cite{vicuna}. Through experimentation, we find that the smallest 7B parameter model is not good enough for our purposes as it fails to follow prompt instructions adequately. In particular, we have found that the model fails to produce short descriptions and often hallucinates either descriptions that were not provided or goes into an effort to describe further information about a person that is not part of their appearance description. The 13B parameter model is able to follow directions more precisely and is adequate for the task at hand. Interestingly, we have found that because the 7B parameter model fails to provide short descriptions, its runtime can exceed that of the 13B model due to needless long generations.  

We follow the process below to obtain sentence descriptions using the LLM. For each sample, we collect the bag-of-words description for all features discussed in sec. \ref{sec:feature_extraction}. 
For all features, we distinguish between {\bf F1} level features, which are primary descriptors such as age, gender and ethnicity and {\bf F2} features, which are more ephemeral, e.g., wearing a hat or beard. All bag-of-features are randomly permuted prior to feeding to the LLM to vary the order they appear in the textual outputs.
For age, we either provide the predicted age by adding some noise (plus minus 1/15 of this value), define an age bracket (plus minus 5 years) or define some descriptive categories for each age (\textsl{``baby'', ``toddler'', ``preschooler'', ``child'', ``teenager'', ``young adult'', ``adult'', ``middle-aged adult'', ``senior adult'', ``elderly''}) with equal probability. Finally, we have noticed a high degree of correlation among detected features. For example, in FFHQ, more than $90\%$ of images classified as attractive using \cite{farl} are women, and among these, more than $80\%$  are classified as wearing makeup. To avoid our data being biased to relate attractiveness with wearing makeup, we randomly drop the "attractive" label when it coincides with makeup with $80\%$ probability. 

We use the following prompt to obtain textual descriptions with Vicuna 13B: 

{\it 
    ``Without elaborating, describe a person with all of the following characteristics: {\bf F1}. They have the following attributes: {\bf F2}. Combine specific characteristics to produce a coherent description. You are encouraged to use synonyms for the provided attributes but not to add information other than what is provided. It is important to use all provided characteristics. Do not repeat characteristics that are provided more than once. Do not repeat these instructions.''
}

Optionally, we append the {\it "The image is blurry"} or {\it "The image is black and white"} in case we have a blurry or a monochrome image. For each image, we obtain more than one LLM output, which we randomly sample from during training, to emphasize the multiplicity of descriptions corresponding to a facial image.

\section{Experiments}\label{sec:experiments}
In this section we train a text-to-image generation model on the images and captions generated following the process presented in sec \ref{sec:method}. We initialize model parameters from Stable Diffusion 2.1 \cite{latent_diffusion, sd21} which we finetune using LoRA \cite{lora}. In particular we train LoRAs attached to the UNet denoiser and the CLIP text encoder of SD2.1.

\subsection{Datasets}
We apply our caption generation pipeline to three publicly available datasets: EasyPortrait \cite{easyportrait}, FFHQ \cite{ffhq} and LAION-Face \cite{farl}. For LAION-Face we apply filtering with the face detector and CLIP as discussed in sec \ref{sec:method} to exclude non-human images as well as low resolution images less than $250 \times 250$ face region. For EasyPortrait and FFHQ we retain all samples. This process results in 40k, 70k and 156k images (and generated captions) for EasyPortrait, FFHQ and LAION-Face respectively.

\subsection{Implementation details}
We apply LoRAs on the UNet denoiser and the CLIP text encoder of SD2.1. We finetune for 30k steps at $768 \times 768$ pixels reesolution using a constant learning rate 1E-5 and $\times 8$ Nvidia V100 cards, each loaded with four samples per training step (total batch size 32). The finetuning process takes approximately two days to complete.

\begin{figure*}[t]
   \centering
   \includegraphics[trim={0 0 0 0}, clip, width=0.995\linewidth]{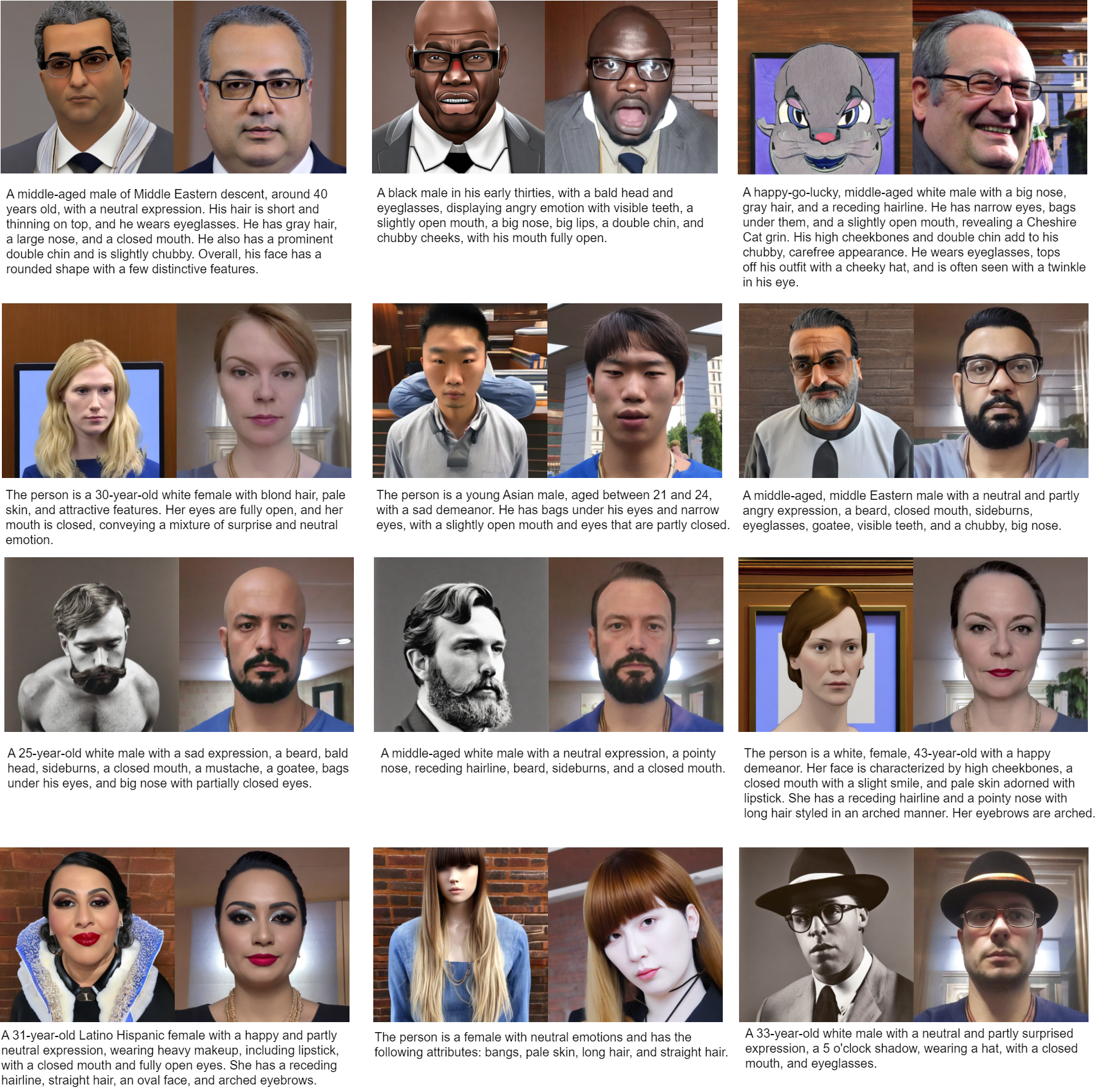}
   \caption{Comparison with SD2.1 base model. For each prompt we show the image generated by the SD2.1 base model (left) as well as our finetuned LoRA model (right).}
    \label{fig:vs_sd21}
\end{figure*}

\begin{figure*}[t]
   \centering
   \includegraphics[trim={0 0 0 0}, clip, width=0.995\linewidth]{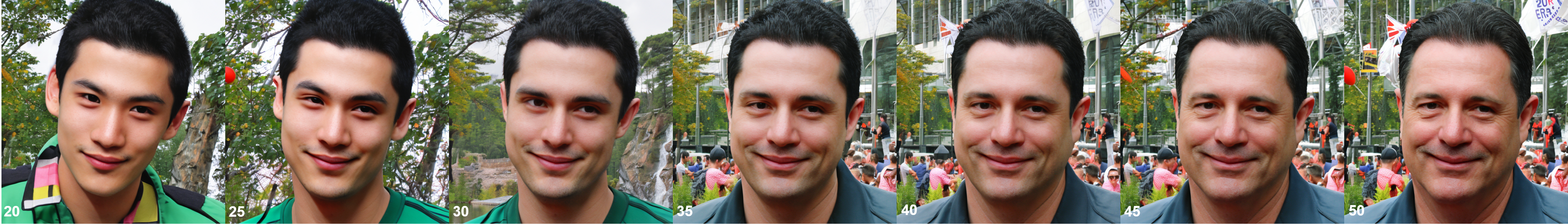}
   \caption{Generated images varying the person's age. Images are generated through the prompt "A $\{$age$\}$ year old white male with black hair and happy expression." where age is substituted by the number shown in each image. We observe that identity characteristics are
   decoupled from age.}
    \label{fig:age}
\end{figure*}

\begin{figure*}[t]
   \centering
   \includegraphics[trim={0 0 0 0}, clip, width=0.995\linewidth]{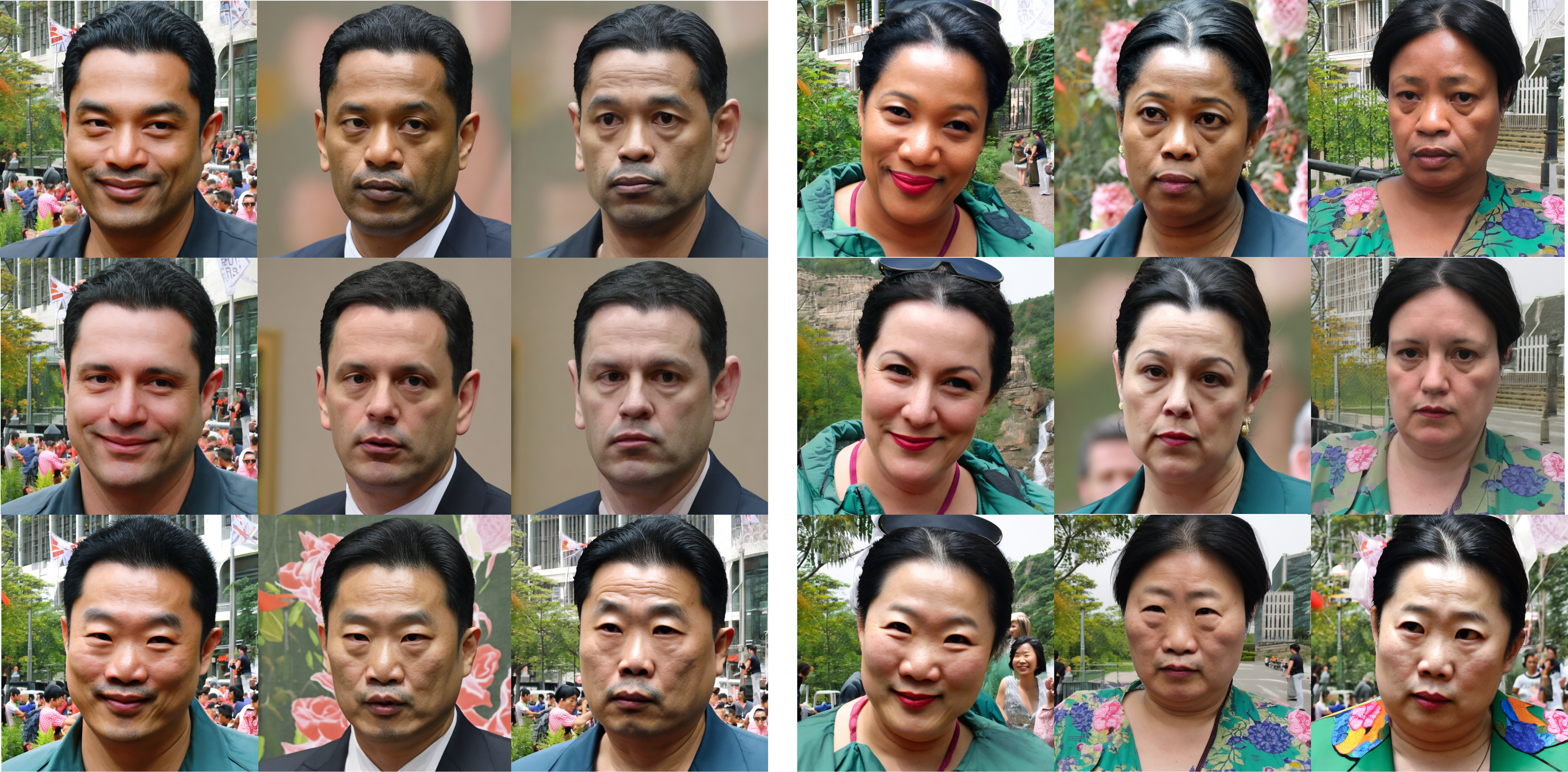}
   \caption{Generated images varying a person's ethnicity, emotion and gender. Images are generated through the prompt "A 40 year old $\{$ethnicity$\}$ $\{$gender$\}$ with black hair and $\{$expression$\}$ expression." where ethnicity, gender and expression are modified accordingly. Gender varies across the left and right figures. Rows represent Black, White and Asian race. Respective figure columns show generations for happy, neutral and sad expressions.}
    \label{fig:gender_ethn_expr}
\end{figure*}

\subsection{Analysis}

In Fig.\ref{fig:vs_sd21} we present a {\bf comparison with the base SD2.1 text-to-image model}. Based on this figure we can make two observations: 1) the finetuned model clearly outputs realistic face images in comparison with the baseline which often includes a cartoonish effect. It is possible to constrain the base model to produce realistic images with prompt engineering, however, this takes considrable effort and computing resourses to generate multiple images varrying the prompt and seed and does not guarantee realistic results. 2) The finetuned models typically follows the prompt more closely, especially when the propmt is complicated and contains multiple instructions.

In Fig.\ref{fig:age} we use the same prompt and random seed and vary the person's age from age 20 to 50, at 5 year intervals. We observe that the identity of the person is preserved to some degree for different ages. More specifically, we note a clustering of identity characteristics around lower (20-25) and higher ($>$35) age values with images in between serving as a transition between the two.   

Similarly, in Fig.\ref{fig:gender_ethn_expr} we vary the gender, ethnicity and emotion of a person. Again, identity characteristics are preserved to some degree between these attributes, altough not as clearly as for age.

\subsection{Limitations}
Since the pipeline involves the use of pretrained face analysis models, their outputs will unavoidably contain all biases found in the employed models.
Similarly, the trained model is finetuned from SD2.1 and as such, inherits all the limitations and biases associated with the base model. 
Overall, biases may manifest in skewed representations across different ethnicities, emotions, age and genders due to the nature of the training data originally used for either the facial analysis models or Stable Diffusion 2.1. Specific limitations include:
\begin{itemize}
    \item {\bf Ethnic and gender biases}: The model may generate images that do not equally represent the diversity of human features in different ethnic and gender groups, potentially reinforcing or exacerbating existing stereotypes.
    \item {\bf Selection bias in finetuning datasets}: The datasets used for finetuning this model were selected with specific criteria in mind, which may not encompass a wide enough variety of data points to correct for the inherited biases of the base model.
    \item {\bf Caption generation bias}: The synthetic annotations used to finetune this model were generated by automated face analysis models, which themselves may be biased. This could lead to inaccuracies in facial feature interpretation and representation, particularly for less-represented demographics in the training data.

\end{itemize}

\section{Conclusion}

We presented a training-free pipeline for the automatic extraction of facial image descriptions to generate large scale image-caption pairs for facial images. These data can be used in training vision-language models for humans. In that direction we used the generated captions to train a text-to-image diffusion model that generates realistic images of humans following prompt conditioning more closely than the base model. The generated captions and finetuned model are made publicly available to encourage further research into this space.

{
    \small
    \bibliographystyle{ieeenat_fullname}
    \bibliography{main}
}


\end{document}